\documentclass[10pt,journal,letterpaper]{ieeetran}

\usepackage{amsmath}
\usepackage{amssymb}
\usepackage{graphicx}
\usepackage{subcaption}
\usepackage{xcolor}
\usepackage{url}
\usepackage{xspace}
\usepackage{hyperref}

\newcommand{\secref}[1]{Sec.~\ref{#1}}
\newcommand{\figref}[1]{Fig.~\ref{#1}}
\newcommand{\tabref}[1]{Tab.~\ref{#1}}

\newcommand{\ProjectName}{CAVEAT\xspace}

\title{\textit{CAVEAT}: Recurrent Multimodal Diffusion Planning for Mapless Aerial Exploration}

\author{Steven Visch$^{1}$ \and
 Nicol\`o Botteghi$^{2}$ \and 
 Antonio Franchi$^{1,3}$ \and
 Barbara Bazzana$^{1}$
\thanks{$^{1}$Robotics and Mechatronics group, Faculty of Electrical Engineering, Mathematics and Computer Science, University of Twente, 7500AE Enschede, The Netherlands.}
    \thanks{$^{2}$Modelling and Scientific Computing Laboratory (MOX), Politecnico di Milano.}
   	\thanks{$^{3}$Department of Computer, Control and Management Engineering, Sapienza University of Rome, 00185 Rome, Italy. }
    \thanks{Email: \texttt{\url{s.l.visch.01@saxion.nl}}, \texttt{\url{nicolo.botteghi@polimi.it}}, \texttt{\url{schol@r-franchi.eu}}, and \texttt{\url{b.bazzana@utwente.nl}}. \par This work was partially funded by the EU: AUTOASSESS project, EU Grant agreement ID: 101120732}
}
\begin{document}

\maketitle


\begin{abstract}
Can exploratory UAV waypoint sequences be generated from multimodal onboard observations and a fixed-dimensional recurrent internal state without maintaining a persistent global map in the deployed policy? We investigate this question through CAVEAT, a diffusion policy conditioned on a recurrent internal state updated from fused LiDAR, visual, and pose features and trained from trajectories generated by the map-based FUELv2 expert. Rolling inference partially warm-starts consecutive predictions, while a temporary local signed distance field provides heuristic obstacle guidance. Simulation results evaluate both inference mechanisms and compare CAVEAT with its demonstration-generating expert. Proof-of-concept experiments on a Flyability Elios~3 demonstrate partial exploration of a previously unseen indoor environment and target-directed visual servoing using a separately trained policy.
\end{abstract}


\section{Introduction}
\label{sec:intro}

Aerial robots are increasingly used to inspect hazardous and difficult-to-access infrastructure while reducing human exposure to these environments \cite{ollero2025tfr}. Such missions frequently take place in initially unknown or incompletely mapped spaces, where planned motions must account for the information available through onboard sensing \cite{bartolomei2020perception}. Exploration then requires the generation of motions that progressively expose previously unobserved regions. This work addresses the high-level generation of UAV waypoint sequences for this purpose.

Established exploration methods support this decision process through explicit spatial representations. Frontier-based methods direct the robot toward boundaries between known and unknown space \cite{yamauchi1997}, whereas next-best-view planners evaluate candidate viewpoints according to their expected information gain \cite{bircher2016nbvp}. Hierarchical frameworks such as FUEL organize local and global planning through incrementally maintained frontiers \cite{fuel}, while coverage-guided methods such as FALCON incorporate global coverage structures \cite{zhang2025tro}. Although their decision mechanisms differ, these approaches represent the environment explicitly through volumetric maps, frontiers, trees, graphs, or coverage structures.

\begin{figure}[t]
    \centering
    \includegraphics[width=\columnwidth]
    {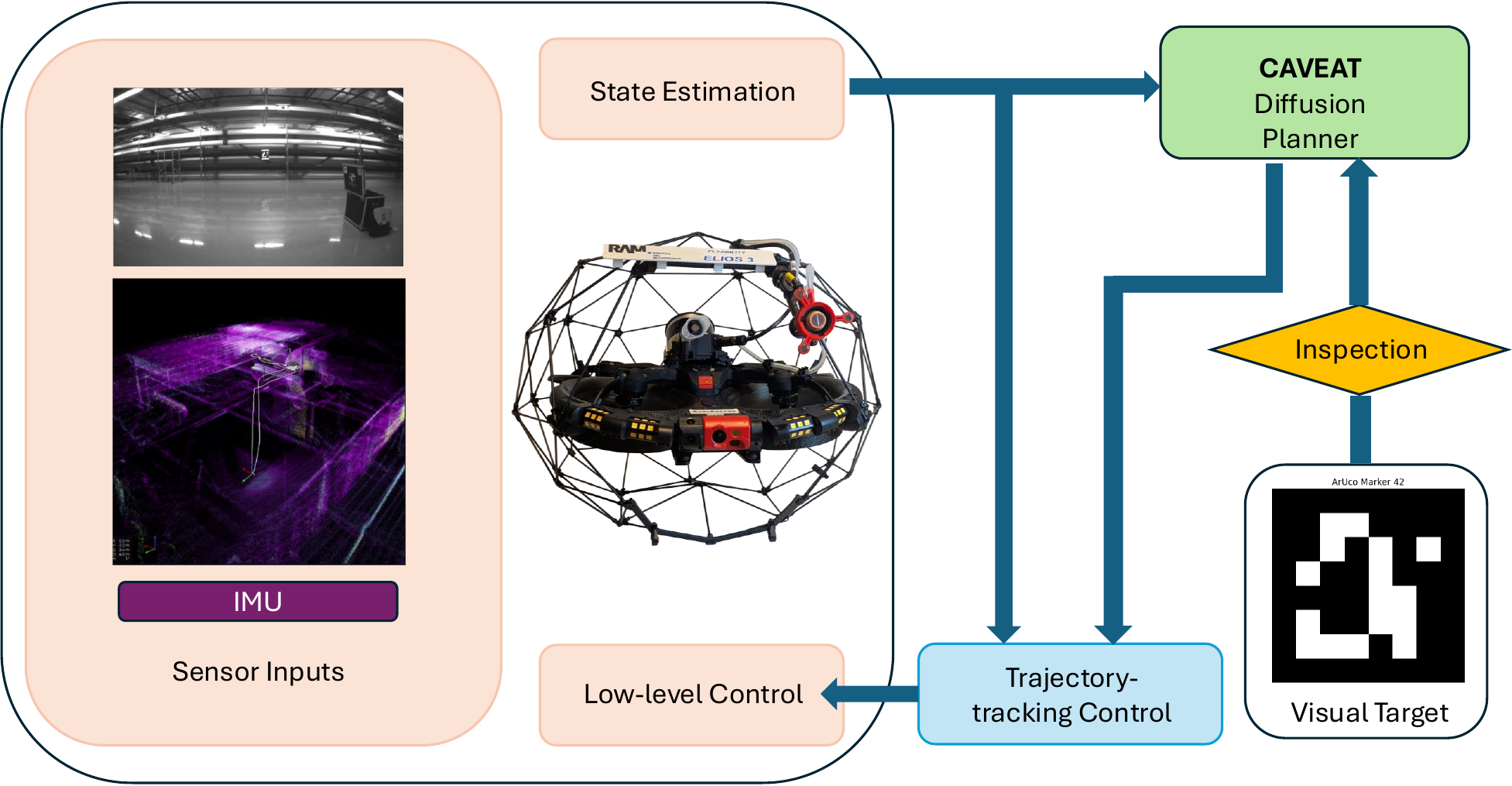}
    \caption{Overview of \ProjectName. Multimodal observations update a recurrent internal state that conditions diffusion-based waypoint generation. Rolling inference partially warm-starts consecutive predictions, while a temporary local SDF provides heuristic obstacle guidance.}
    \label{fig:method_overview}
\end{figure}

Learning-based exploration offers a complementary representation of the decision process. Imitation learning can transfer trajectories generated by an exploration expert to a policy operating on sensor observations \cite{reinhart2020icra}, while learning-based informative path planning more broadly studies how task-relevant representations and decision rules can be inferred from data \cite{popovic2024ras}. A learned policy can thereby generate exploration actions without executing the complete demonstration-generating pipeline at deployment. This possibility raises the central question considered here: can the spatial and historical information required for useful exploration be represented by a compact recurrent internal state rather than by a persistent global spatial representation inside the deployed policy?

To investigate this question, we introduce \emph{CAVEAT}, a \emph{C}onditional diffusion policy for \emph{A}utonomous LiDAR-\emph{V}isual-Inertial \emph{E}xploration using \emph{A}erial robo\emph{T}s. Diffusion models learn conditional distributions over ordered action or trajectory sequences \cite{ddpm} and have been applied to robot motion planning and visuomotor policies \cite{carvalho2023motion,chi2025diffusion}. These properties make diffusion models suitable for generating complete waypoint sequences and representing potentially multimodal action distributions. Their application to exploration requires a conditioning representation containing the current observations and relevant information from the observation history.

Diffusion-based exploration has already been investigated in DARE and GUIDE, which condition action generation on structured representations of observed and inferred environmental information \cite{cao2025icra,che2026arxiv}. \ProjectName studies a different conditioning architecture. Modality-specific encoders process onboard LiDAR, visual, and pose observations, while a gated recurrent unit (GRU) compresses the sequence of fused features into a fixed-dimensional recurrent internal state \cite{GRU}. This state conditions exploratory waypoint generation without requiring the learned policy to maintain a persistent global map or graph at deployment. The hypothesis examined here is that the recurrent state can retain sufficient task-relevant information from the observation history to support useful exploratory motion.

As summarized in \figref{fig:method_overview}, \ProjectName generates four-dimensional waypoint sequences from the recurrent internal state. The policy is trained by imitation from trajectories produced by FUELv2, a mature, open-source, map-based exploration framework compatible with the sensing assumptions considered here \cite{fuelv2}. The expert uses an explicit global spatial representation to generate demonstrations, while the deployed policy acts from current multimodal observations and its recursively updated internal state.

Two inference mechanisms support receding-horizon deployment. Rolling inference partially warm-starts each denoising process from the preceding prediction by retaining its initial portion and progressively perturbing later waypoints \cite{diffuse_cloc}. This mechanism carries information from the previous predicted trajectory, whereas the GRU state summarizes encoded observation history. In parallel, a temporary two-dimensional signed distance field (SDF) constructed from the current LiDAR scan guides nearby waypoints toward greater obstacle clearance \cite{oleynikova2016rss}. The resulting sparse waypoints are (i) interpolated and passed to an existing NMPC tracking layer \cite{vision_mpc} for the training and simulation runs, or (ii) passed to a polynomial-based trajectory generation and tracking module for the real-world Elios~3 experiments.

The contribution of this work is an empirical investigation of recurrent multimodal conditioning for diffusion-based aerial exploration, together with rolling waypoint initialization and local SDF guidance for deployment. The evaluation uses FUELv2 demonstrations collected in four simulated environment families, studies the two inference mechanisms, and compares \ProjectName with its demonstration-generating expert. Proof-of-concept experiments deploy the policy on a Flyability Elios~3 in a previously unseen indoor environment. A separately trained instance of the same architecture further demonstrates target-directed visual servoing as an inspection-oriented secondary behavior.

\section{Method}
\label{sec:main}

\ProjectName is a receding-horizon policy for high-level UAV waypoint generation. At decision step $t$, it processes the current LiDAR point cloud, forward-facing camera image, and visual-inertial pose estimate together with its recurrent internal state. The policy generates a sequence of four-dimensional waypoints expressed in the odometry frame, executes the first waypoint, and replans from updated observations. The sparse waypoints are interpolated subject to velocity, acceleration, angular-velocity, and angular-acceleration constraints before being passed to the trajectory tracking controller: ~\cite{Bicego_2020,vision_mpc} for training and simulation, and ~\cite{richter2016polynomial} + PID control for real-world deployment. This section focuses on the exploration policy; a separately trained instance of the same architecture is evaluated for target-directed visual servoing in \secref{subsec:real_world_servoing}.

\subsection{Recurrent Multimodal Diffusion Policy}
\label{subsec:diff_policy}

\begin{figure}[t]
    \centering
    \includegraphics[width=0.99\columnwidth]
    {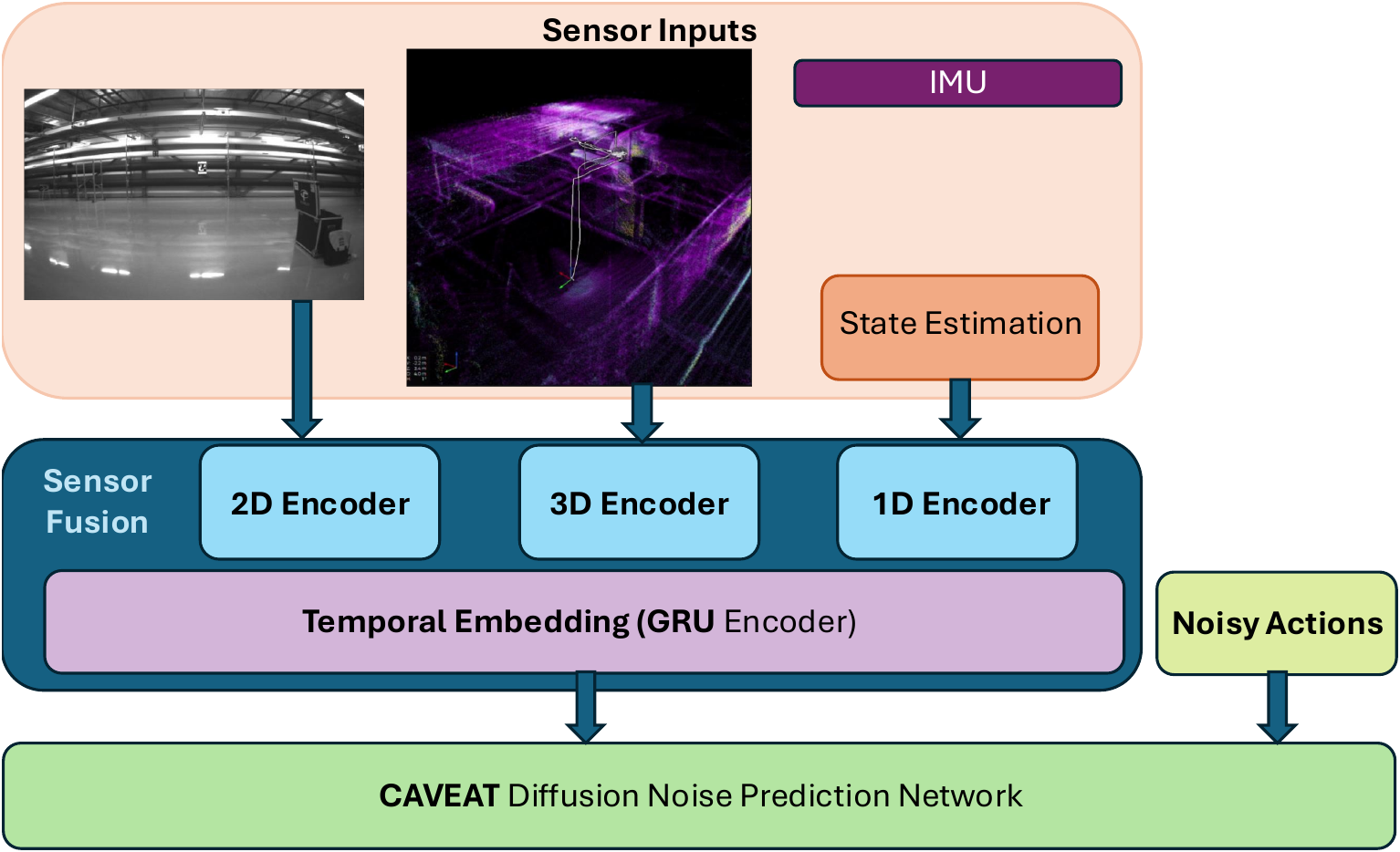}
    \caption{Architecture of the \ProjectName exploration policy. Modality-specific encoders process the current LiDAR, visual, and pose observations. Their fused features update a recurrent internal state that conditions diffusion-based waypoint generation through feature-wise linear modulation.}
    \label{fig:method_low_level}
\end{figure}

As illustrated in \figref{fig:method_low_level}, the exploration policy comprises modality-specific observation encoders, a feature-fusion network, a gated recurrent unit (GRU), and a conditional diffusion model. The encoders extract compact features from the current observations, and the GRU recursively updates a fixed-dimensional internal state representing their history. This state conditions the diffusion model that generates the next waypoint sequence.

\subsubsection{Multimodal observation encoding}
\label{subsec:sensor_fusion}

The 3D encoder processes the LiDAR point-cloud after farthest-point sampling to 1024 points. Following the permutation-invariant design adopted in \cite{3d_diffusion}, shared pointwise transformations and symmetric max pooling produce a 512-dimensional feature vector. Sampling bounds the encoder input size, while symmetric pooling makes its output independent of point ordering.

The visual encoder is a compact three-stage ResNet-style network inspired by \cite{semanticRL}. Each stage comprises max-pooling downsampling followed by a residual block. Adaptive average pooling reduces the resulting feature map to an $8\times8$ spatial grid, after which a linear projection with ELU activation produces a 512-dimensional feature vector. The coarse spatial grid preserves approximate image-location information, which is particularly relevant to the separately trained target-directed behavior. Group normalization is used to accommodate the small or variable batch sizes employed during training.

The seven-dimensional position estimate in the Visual Inertial Odometry (VIO) fixed local frame, comprising the 3D position and the quaternion orientation, is processed by a two-layer MLP with ELU activations, producing a 128-dimensional feature vector. The modality-specific features are concatenated and compressed by a two-layer MLP into the fused observation feature
\(
    \mathbf{z}_t \in \mathbb{R}^{n_z},
\)
where $n_z=512$. This learned fusion determines the relative contribution of the sensing modalities from the training data.

\subsubsection{Recurrent internal state}
\label{subsec:recurrent_state}

A single observation provides only a local and instantaneous description of the environment. To retain task-relevant information from preceding observations without maintaining a persistent global map, the fused feature recursively updates a GRU \cite{GRU}:
\begin{equation}
    \mathbf{h}_t
    =
    \operatorname{GRU}_{\theta}
    \left(\mathbf{z}_t,\mathbf{h}_{t-1}\right),
    \qquad
    \mathbf{h}_t\in\mathbb{R}^{n_h},
    \label{eq:gru_update}
\end{equation}
where $\mathbf{h}_t$ is the recurrent internal state and
$n_h=512$. Its dimension is independent of the episode duration. The state is learned for waypoint prediction and is neither constructed nor supervised as a geometric map. At the beginning of each episode, it is initialized to zero for each processed observation sequence, whose length is 33.

\subsubsection{Conditional waypoint diffusion}
\label{subsec:conditional_diffusion}

The noise-prediction network follows the one-dimensional temporal CNN architecture of \cite{chi2025diffusion}. Its convolutions operate along the ordered waypoint sequence, providing an inductive bias toward local dependence between adjacent waypoints. The recurrent state $\mathbf{h}_t$ conditions the intermediate network features through feature-wise linear modulation (FiLM) \cite{FiLM}.

Let $\mathbf{X}_t^{(k)}$ denote the noisy waypoint sequence at diffusion step $k$. Starting from an initially noisy sequence, iterative noise prediction conditioned on $\mathbf{h}_t$ yields
\[
    \mathbf{X}_t
    =
    \left(
    \mathbf{x}_{t|t},
    \ldots,
    \mathbf{x}_{t+H-1|t}
    \right),
    \qquad
    \mathbf{x}_{j|t}\in\mathbb{R}^{4},
\]
where $H=8$ and
\[
    \mathbf{x}_{j|t}
    = [x_j, y_j, z_j, \theta_j].
\]
Here, $(x_j,y_j,z_j)$ denote the absolute 3-D waypoint position in metres in the simulation/world frame, and $\theta_j$ denotes the absolute yaw angle in radians.
Only $\mathbf{x}_{t|t}$ is executed before updated observations are processed. The observation horizon is 33 observations (32 history steps plus the current observation). The diffusion model is trained with 100 diffusion steps, with a squared-cosine noise schedule with variance ranging from \(10^{-4}\) to \(0.02\) and fixed-small variance, while the network is trained to predict the added noise. FiLM conditioning is enabled in every conditional residual block of the 1-D U-Net, using the fused observation feature together with the diffusion-step embedding to predict per-channel scale and bias.

\subsection{Rolling Inference and Local Obstacle Guidance}
\label{subsec:guided_inference}

The recurrent internal state summarizes the observation history but does not directly constrain stochastic waypoint sequences generated at consecutive decision steps to agree. Independent initialization may therefore produce abrupt changes in the planned direction. \ProjectName addresses consecutive-plan consistency through rolling initialization, which reuses part of the preceding waypoint prediction. Separately, a temporary local signed distance field (SDF) corrects waypoints near currently observed obstacles. Rolling initialization, the local SDF, and the GRU state respectively represent previous-plan information, nearby geometry, and encoded observation history.

\subsubsection{Rolling inference}
\label{subsec:rolling_inference}

At the first decision step, the diffusion process is initialized from Gaussian noise. At each subsequent step, the preceding waypoint sequence is shifted after its first waypoint has been executed. The first $N_{\mathrm{fixed}}$ remaining waypoints are retained, while progressively increasing Gaussian noise is applied to later waypoints; the final waypoint is fully resampled. Inspired by \cite{diffuse_cloc}, this partially perturbed sequence initializes the next denoising process, which is conditioned on the updated recurrent state:
\begin{equation}
\widetilde{\mathbf{x}}_{t+j|t}
    =
    \begin{cases}
    \mathbf{x}_{t+j|t-1}\qquad\mathrm{if}\,j\leq N_{fixed}
    \\
    \mathbf{x}_{t+j|t-1}
    +
    \alpha_j\,
    \boldsymbol{\epsilon}_j,
    \qquad
    \boldsymbol{\epsilon}_j
    \sim
    \mathcal{N}(\mathbf{0},\mathbf{I}),\,\mathrm{else},    
    \end{cases}
    \label{eq:rolling_initialization}
\end{equation}
where $N_{fixed}=3$, and the noise amplitude $\alpha_j$ schedule increases linearly from 0 to 1, namely $\alpha_j\in
[0.1;0.325;0.55;0.775;1.0]$, with the final waypoint replaced by an independently sampled Gaussian-noise vector. 

\subsubsection{Local SDF guidance}
\label{subsec:obstacle_avoidance}

A diffusion policy does not inherently constrain its waypoint predictions to be collision-free. At each decision step, \ProjectName constructs a temporary two-dimensional signed distance field (SDF) from the current LiDAR point cloud \cite{oleynikova2016rss}. Points within $\pm0.5\,\mathrm{m}$ of the current flight altitude are projected onto the horizontal plane to approximate nearby planar geometry. The resulting SDF is local to the current observation and is not accumulated into a persistent global map.

Let $\mathbf{p}=\Pi\mathbf{x}\in\mathbb{R}^{2}$ denote the planar-position components of a normalized four-dimensional waypoint $\mathbf{x}$, where $\Pi=
\begin{bmatrix}
1&0&0&0\\
0&1&0&0
\end{bmatrix}$. During the final four denoising steps, the planar position of each waypoint whose SDF value is below the clearance threshold $d_{\min}=0.5\,\mathrm{m}$ is adjusted according to:
\begin{equation}
\mathbf{p}_{\mathrm{adjusted}}
=
\begin{cases}
\mathbf{p}
+
\lambda\nabla_{\mathbf{p}}\operatorname{SDF}(\mathbf{p}),
&
\operatorname{SDF}(\mathbf{p})<d_{\min},
\\
\mathbf{p},
&
\text{otherwise},
\end{cases}
\label{eq:sdf_guidance}
\end{equation}
while the remaining waypoint components are left unchanged. $\lambda$ is scaled according to the corresponding action-normalization scales, and the resulting correction is applied with a factor of one half. This SDF-gradient correction does not modify the vertical position and yaw. In addition to the local clearance test, CAVEAT checks whether the straight-line segment from the current UAV position to each predicted waypoint is blocked. The segment is sampled at intervals of half the SDF grid resolution, i.e., $0.025\,\mathrm{m}$. A waypoint is considered blocked if the segment leaves the local SDF grid, intersects an occupied cell with non-positive SDF, or passes through a region whose SDF value remains at or below the $0.2\,\mathrm{m}$ robot-clearance radius for a distance of at least twice that radius. For a blocked waypoint, the planar waypoint position is instead displaced toward the current UAV position, with the displacement scaled by the action-normalization factor.

The guidance is consequently a local heuristic rather than a hard collision constraint. It uses only the currently observed LiDAR geometry, does not maintain a persistent map, and does not strictly guarantee collision-free waypoint sequences or collision-free executed trajectories.

\subsection{Expert Demonstrations and Training}
\label{subsec:expert_training}

The exploration policy is trained by imitation from trajectories generated in Gazebo by FUELv2 \cite{fuelv2}. FUELv2 was selected because its open-source ROS implementation supports exploration from range measurements and odometry while accounting for imperfect state estimation. During demonstration generation, FUELv2 maintains a global TSDF map and executes its map-based exploration pipeline. At deployment, FUELv2 is replaced by the learned policy described above.

Direct integration of the simulated sparse LiDAR measurements into the FUELv2 mapping component caused free space in front of observed surfaces to be classified as occupied. Explicit ray-based free-space carving was therefore added during demonstration generation: voxels between the sensor origin and each LiDAR return receive positive signed-distance evidence. This modification is used only to generate the expert demonstrations.

The diffusion model is trained with the standard denoising objective \cite{ddpm}. For each training sequence, the expert waypoint trajectory is independently normalized component-wise using dataset-wide limits. A diffusion timestep is sampled uniformly from the 100 training diffusion steps, Gaussian noise is added according to the squared-cosine noise schedule, and the network is trained to predict the added noise using a mean-squared-error objective.

The modality encoders, fusion network, GRU, and conditional diffusion U-Net are optimized jointly. Training uses the AdamW optimizer with a learning rate of $10^{-4}$, $\beta_1=0.95$, $\beta_2=0.999$, $\epsilon=10^{-8}$, and weight decay $10^{-6}$. A cosine learning-rate schedule is used, with 500 warm-up steps. The training batch size is 96 and training is performed for 50 epochs. An exponential moving average of the model parameters is maintained during training.

Each training sample contains 40 consecutive recorded timesteps. The first 32 timesteps provide the observation history used by the temporal encoder, while the following eight timesteps form the target waypoint sequence for the diffusion model. The resulting training problem therefore maps a history of 33 multimodal observations (32 history steps plus the current observation) to an eight-waypoint expert trajectory. During inference, the same architecture predicts an eight-waypoint sequence, but only the first waypoint is executed before the observation history and recurrent state are updated and the prediction horizon is rolled forward.

The expert waypoint sequence is constructed by taking eight consecutive command samples following the current observation, without interpolation on the demonstration data. The four-dimensional command is stored as three position coordinates together with the yaw value supplied by the command interface. Before diffusion training, each of the four action dimensions is independently normalized to ([-1,1]) using the empirical minimum and maximum values computed from the training dataset.

The sequence sampler operates independently within episode boundaries. At the beginning of an episode, insufficient history is handled by left-padding the sampled sequence with zeros. The planner is currently configured to stop at the end of an episode when it cannot create a full action sequence.

For the target-directed visual-servoing experiment, a separate model with the same architecture is trained on simulated trajectories that approach a predefined visual target from randomized initial conditions. Exploration and target-directed visual servoing therefore use distinct trained policies.

\section{Experimental Evaluation}
\label{sec:exp}

The evaluation addresses three questions: (i) whether \ProjectName generates effective exploratory motion in simulation, (ii) how rolling inference and local SDF guidance affect occupancy coverage per traveled distance, and (iii) whether the learned policy can be deployed on the physical target platform.

\subsection{Experimental Setup and Metrics}
\label{subsec:experimental_setup}

\paragraph{Platform and sensing}
The target platform is the Flyability Elios~3 inspection UAV. The simulated platform reproduces the sensing configuration and odometry-frame convention relevant to the high-level policy; waypoint interpolation and tracking are delegated to the downstream controller. The simulated sensor suite comprises a 360-degree Ouster OS-2 32-beam LiDAR, a forward-facing monochrome visual-inertial camera with a resolution of $640\times400$ pixels, and an additional forward-facing LiDAR. The UAV pose estimate is assumed to be available and is provided by the proprietary onboard estimation stack on the physical Elios~3.

The forward-facing LiDAR is used in simulation to provide structured depth measurements to the FUELv2 demonstration-generation pipeline and to support the local geometric guidance and coverage evaluation. Its field of view also defines the reported forward-looking occupancy coverage, which measures the fraction of the predefined forward-view spatial region that has been observed by the LiDAR. Specifically, the observed spatial region is compared with a reference set representing complete coverage of the corresponding forward-looking field of view, and the coverage is reported as the ratio of observed elements to the total number of elements in this reference set, expressed as a percentage. The Ouster LiDAR is therefore not an input to the learned observation encoder; rather, it is used separately for simulation-side geometric processing, including demonstration generation and local SDF guidance.

\begin{figure}[t]
    \centering
    \begin{subfigure}[b]{0.48\linewidth}
        \centering
        \includegraphics[width=\linewidth,height=2.45cm,keepaspectratio]
        {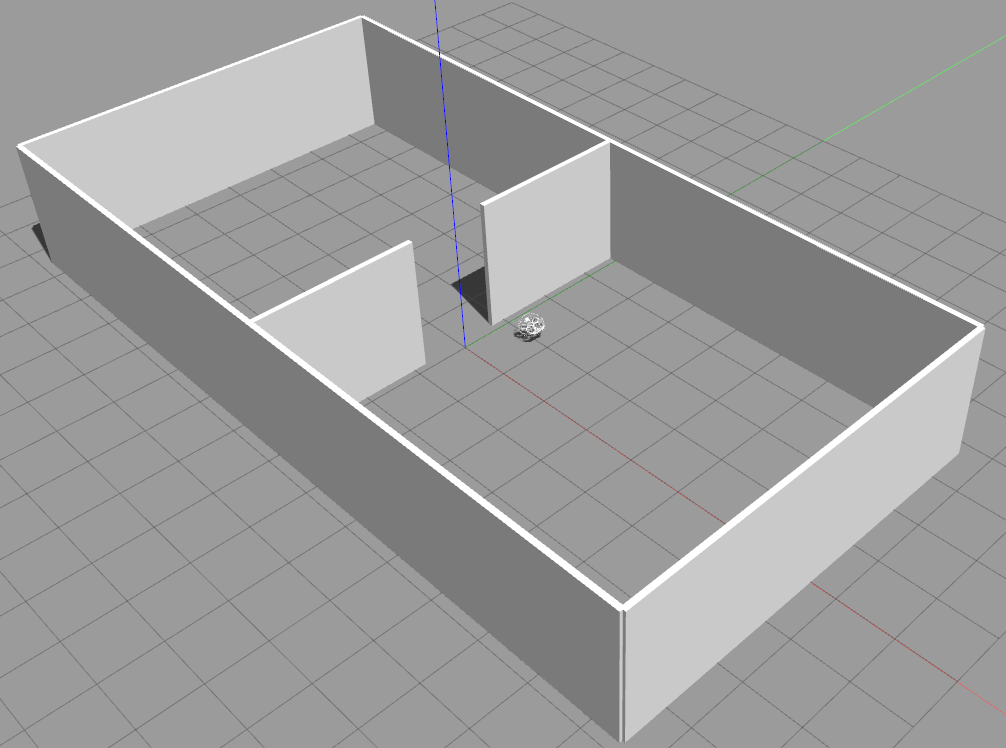}
        \caption{Two rooms.}
    \end{subfigure}
    \hfill
    \begin{subfigure}[b]{0.48\linewidth}
        \centering
        \includegraphics[width=\linewidth,height=2.45cm,keepaspectratio]
        {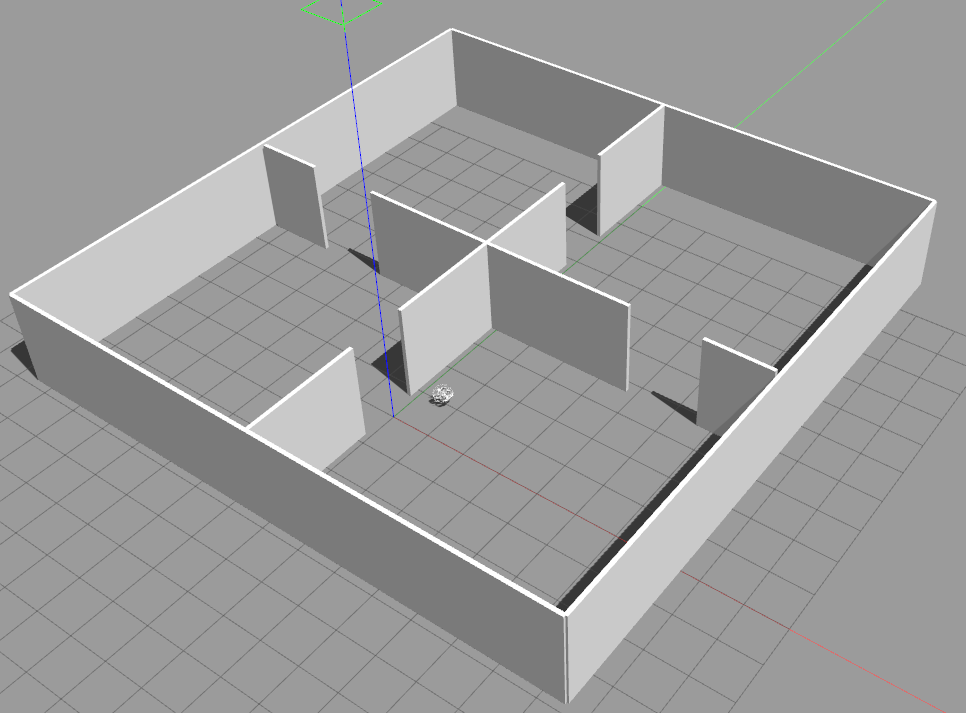}
        \caption{Four rooms.}
    \end{subfigure}

    \vspace{0.10cm}

    \begin{subfigure}[b]{0.48\linewidth}
        \centering
        \includegraphics[width=\linewidth,height=2.45cm,keepaspectratio]
        {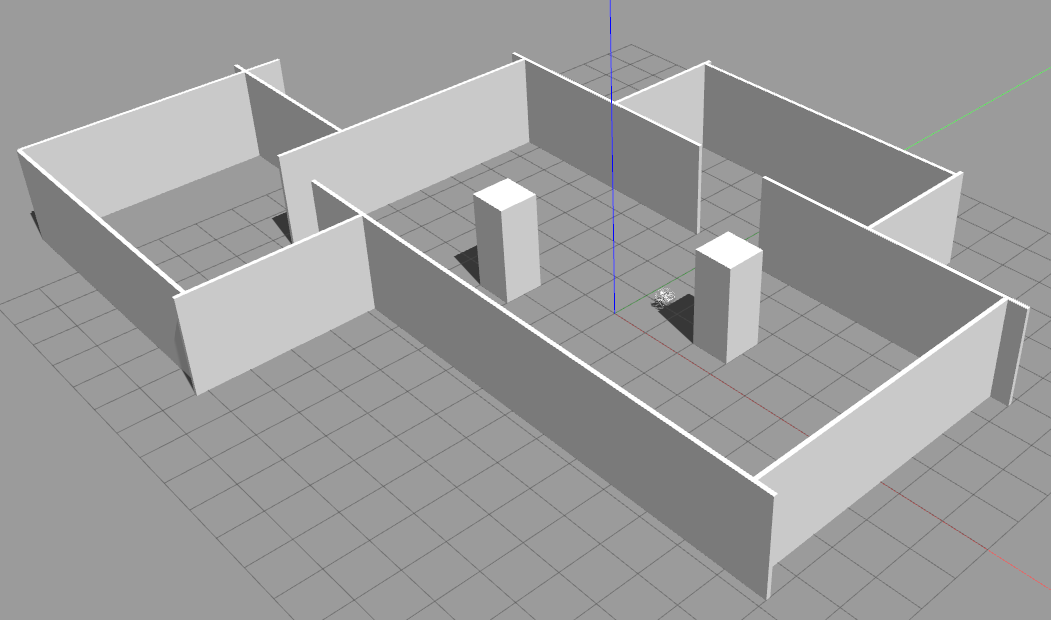}
        \caption{Complex topology.}
        \label{fig:complex_env}
    \end{subfigure}
    \hfill
    \begin{subfigure}[b]{0.48\linewidth}
        \centering
        \includegraphics[width=\linewidth,height=2.45cm,keepaspectratio]
        {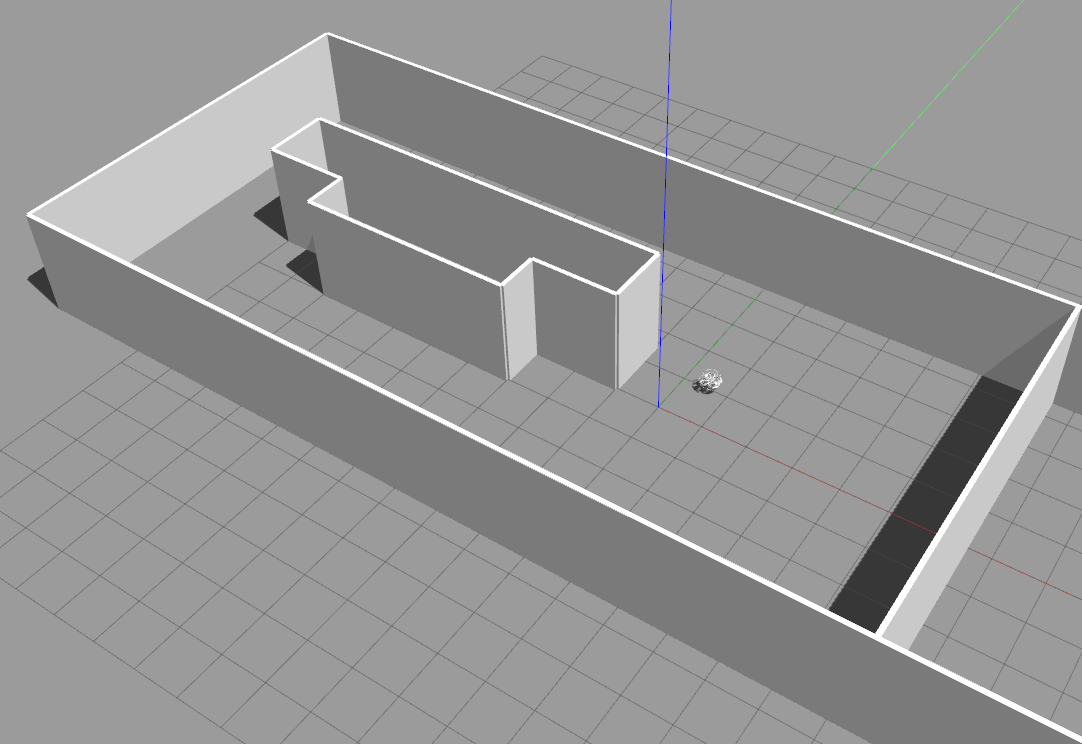}
        \caption{Occluded long room.}
        \label{fig:simple_carre_env}
    \end{subfigure}
    \caption{Gazebo environment families used to collect the expert exploration demonstrations. Environments $A_{\mathrm{train}}$--$D_{\mathrm{train}}$ correspond to panels (a)--(d), respectively.}
    \label{fig:simulation-setup}
\end{figure}

\paragraph{Training environments and demonstrations}
Expert demonstrations were generated in Gazebo using FUELv2 in the four environment families shown in \figref{fig:simulation-setup}. The layouts range from connected rooms to environments containing multiple rooms, occlusions, and more complex connectivity. Each episode starts from a randomized feasible location, where the odometry-frame origin is initialized at floor level consistently with the physical Elios~3.

\begin{figure}[t]
\centering
\begin{subfigure}[c]{0.45\linewidth}
    \centering
    \includegraphics[width=\linewidth]
    {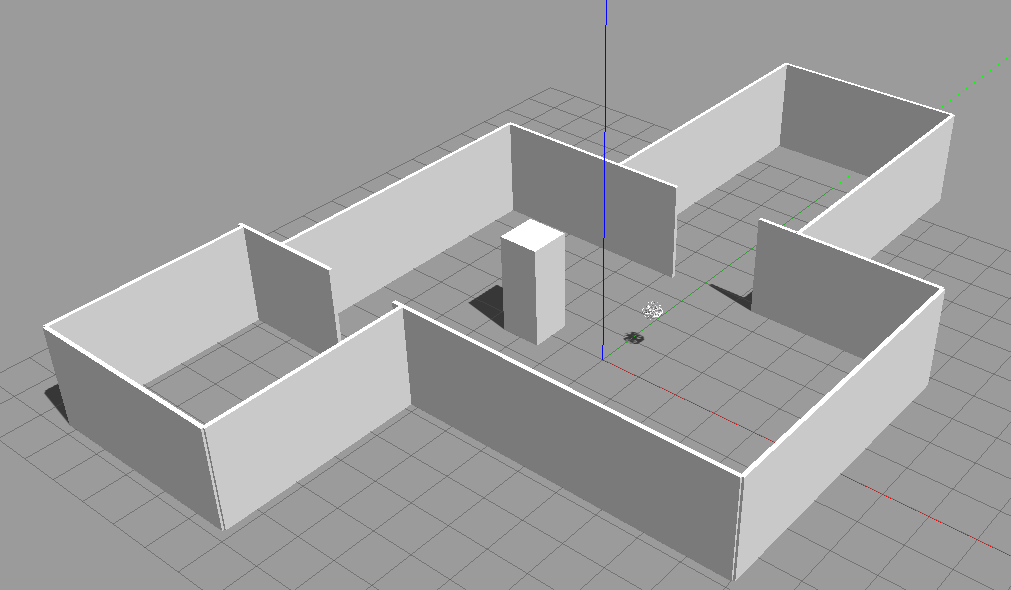}
    \caption{Complex topology.}
\end{subfigure}
\hfill
\begin{subfigure}[c]{0.5\linewidth}
    \centering
    \includegraphics[width=\linewidth]
    {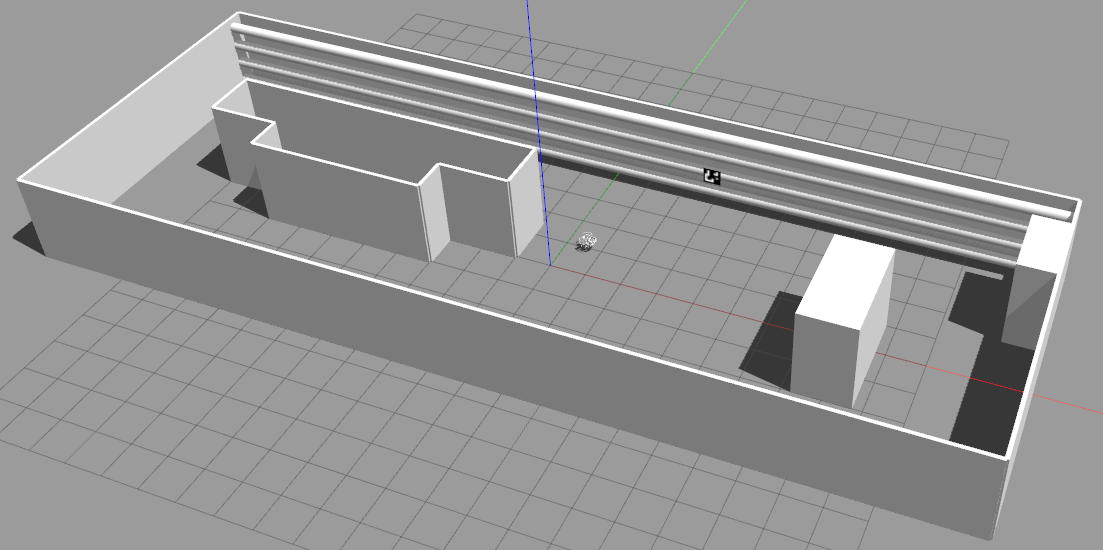}
    \caption{Occluded long room.}
\end{subfigure}
\caption{Gazebo test environments. Environments $A_{\mathrm{test}}$--$B_{\mathrm{test}}$ correspond to panels (a)--(b), respectively.}
\label{fig:real-world-gazebo}
\end{figure}

Demonstrations were recorded at $1\,\mathrm{Hz}$ with a maximum expert-flight speed of $0.5\,\mathrm{m/s}$. Each record contains the UAV pose estimate, the downsampled LiDAR point clouds, the forward-facing camera image, and the expert goal pose used to construct the training waypoint sequence. The complete dataset contains 2,674 trajectories and 199,603 recorded time steps, as summarized in \tabref{tab:dataset-summary}. For policy training, trajectories are randomly divided into training and validation subsets, with $2\%$ of trajectories reserved for validation and the remaining $98\%$ used for training. The split is generated at the trajectory level using a fixed random seed of 42, so samples from a single trajectory are not distributed between the training and validation sets. The evaluation episodes are generated separately from the training/validation dataset and therefore do not constitute held-out samples from the recorded demonstration trajectories. They use the test configuration of the simulated environment rather than replaying the training demonstrations.
\begin{table}[t]
\centering
\setlength{\tabcolsep}{3.6pt}
\renewcommand{\arraystretch}{0.92}
\begin{tabular}{lc@{\hspace{9pt}}lc}
\hline
Trajectory statistic & Steps &
Environment & Trajectories \\
\hline
Mean               & 74.65 & $A_{\mathrm{train}}$     & 1000 \\
Median             & 79.00 & $B_{\mathrm{train}}$     & 700  \\
Standard deviation & 22.64 & $C_{\mathrm{train}}$     & 400  \\
Minimum            & 1     & $D_{\mathrm{train}}$     & 574  \\
Maximum            & 114   & Total & 2674 \\
\hline
\end{tabular}
\caption{Exploration-demonstration dataset, comprising 2,674 trajectories and 199,603 recorded time steps.}
\label{tab:dataset-summary}
\end{table}

\paragraph{Evaluation protocol}
Simulation performance is measured by exploration/occupancy coverage as a function of UAV traveled distance. Each evaluation trial starts from a randomized initial condition in the simulated test environment. The central curve represents the mean performance across trials, while the shaded region represents the corresponding standard deviation across trials. Random seeds are not fixed across evaluation trials: each trial initializes Python, NumPy, and PyTorch random generators using a seed derived from the current system time. Consequently, the evaluation protocol samples different randomized initial conditions across runs rather than providing strictly deterministic, seed-reproducible trials.

Each trial terminates when the frontier set maintained by a parallel FUELv2 evaluation instance becomes empty or when the evaluated planner reaches a local minimum. This evaluation-side map is not provided to \ProjectName.
The policy uses one action step per replanning cycle. If the predicted action does not correspond to a valid neighboring node of the current UAV position, the implementation applies a local collision-avoidance fallback by selecting the neighboring node whose direction is most consistent with the predicted future motion. A collision/invalid-motion event is therefore handled by the evaluation controller rather than being directly executed as an invalid transition. 

Evaluation is performed using PyTorch 2.4.0 with CUDA 12.4 in the supplied environment specification, with inference configured to use a GPU. The test driver launches 10 parallel Ray workers sharing one GPU.

\subsection{Simulation Results}
\label{subsec:simulation_results}

The simulation study evaluates rolling inference, examines the effect of the local SDF clearance threshold, and compares the complete policy with its FUELv2 demonstration-generating expert. All learned-policy parameters are held fixed within each comparison.
\begin{figure}[t]
\centering

\begin{subfigure}[c]{0.48\linewidth}
    \centering
    \includegraphics[width=\linewidth]
    {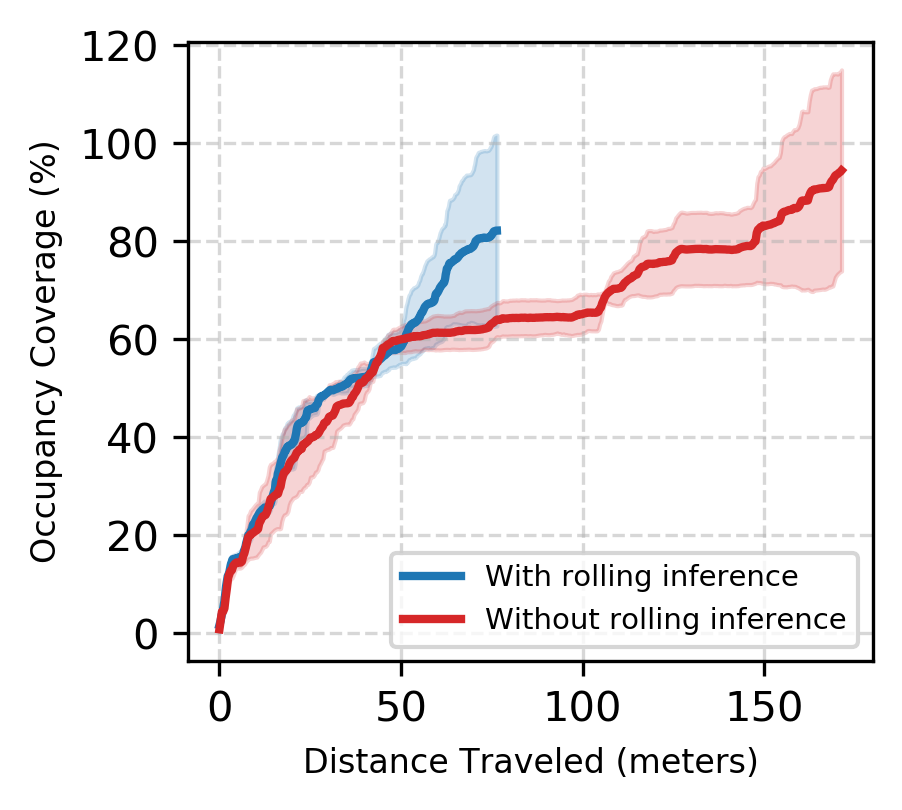}
    \caption{Rolling, env.~$A_{\mathrm{test}}$.}
    \label{fig:rolling-A}
\end{subfigure}
\hfill
\begin{subfigure}[c]{0.48\linewidth}
    \centering
    \includegraphics[width=\linewidth]
    {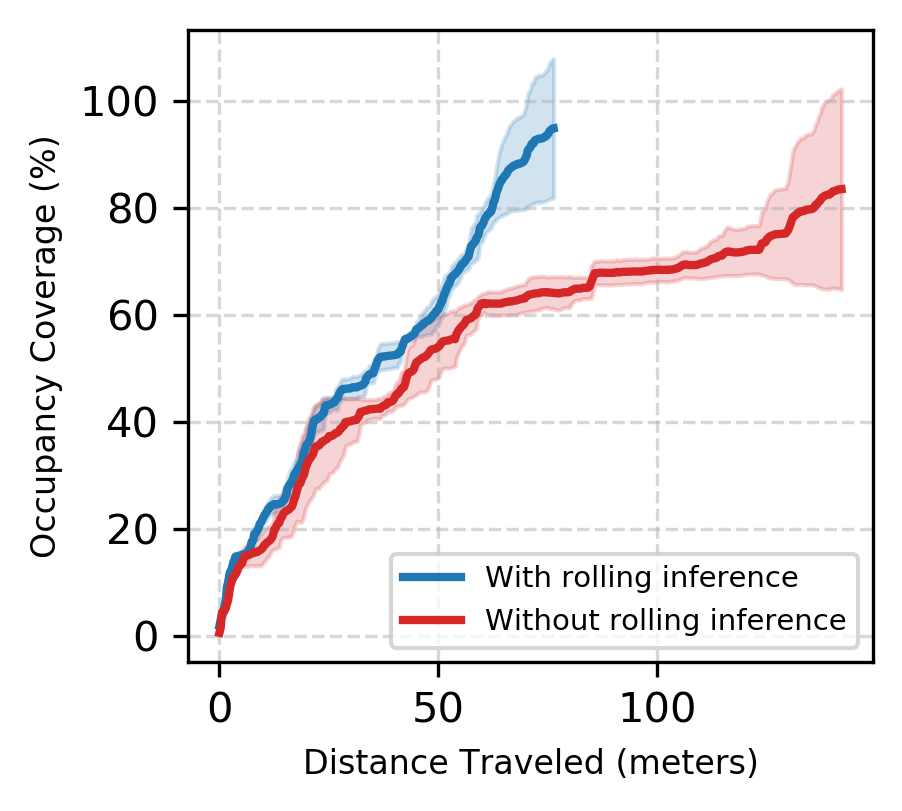}
    \caption{Rolling, env.~$B_{\mathrm{test}}$.}
    \label{fig:rolling-B}
\end{subfigure}

\vspace{0.08cm}

\begin{subfigure}[c]{0.48\linewidth}
    \centering
    \includegraphics[width=\linewidth]
    {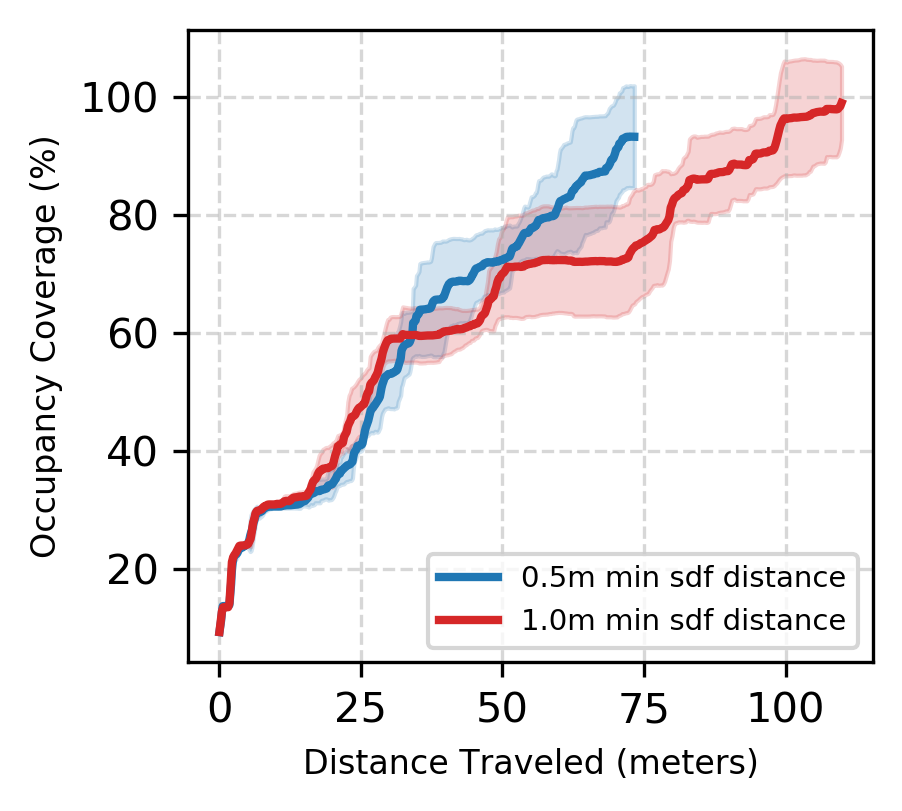}
    \caption{SDF, env.~$A_{\mathrm{test}}$.}
\end{subfigure}
\hfill
\begin{subfigure}[c]{0.48\linewidth}
    \centering
    \includegraphics[width=\linewidth]
    {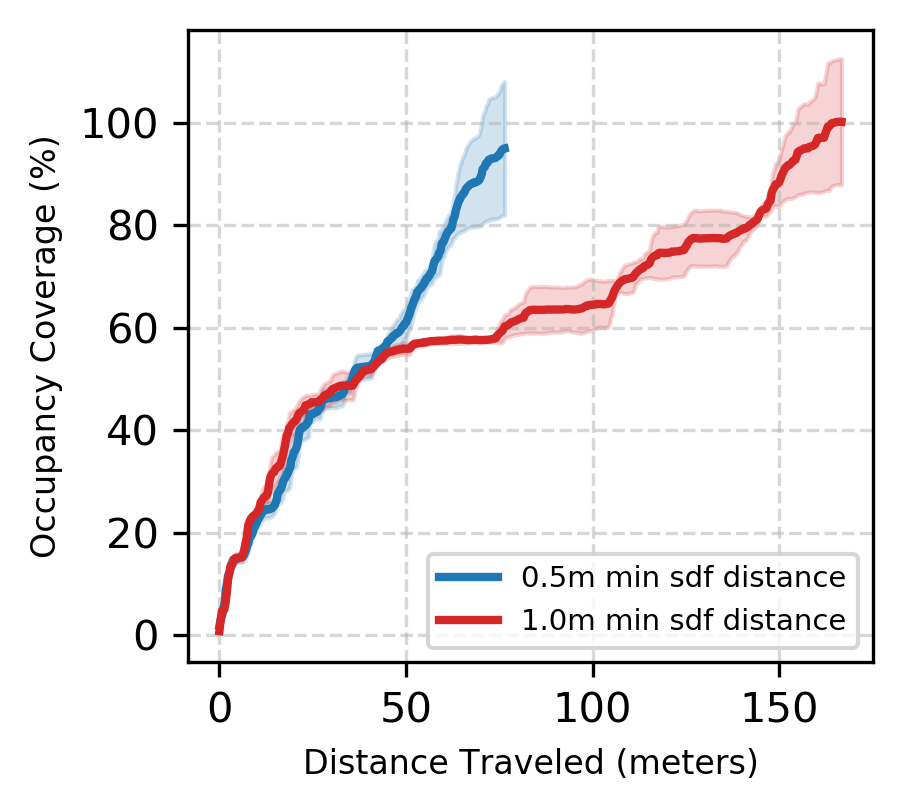}
    \caption{SDF, env.~$B_{\mathrm{test}}$.}
\end{subfigure}
\vspace{0.08cm}

\begin{subfigure}[c]{0.48\linewidth}
    \centering
    \includegraphics[width=\linewidth]
    {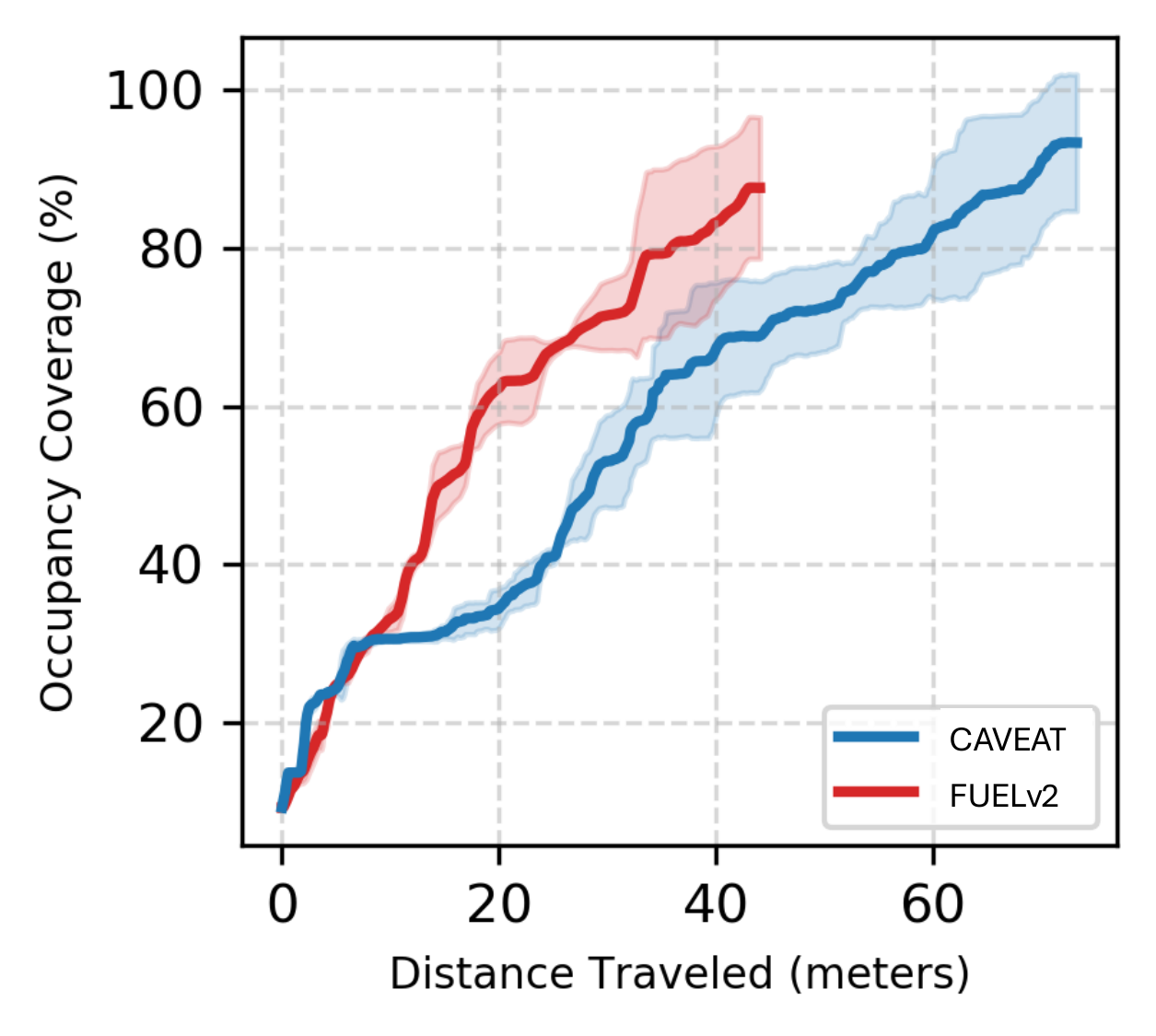}
    \caption{\ProjectName vs. FUELv2, env.~$A_{\mathrm{test}}$.}
\end{subfigure}
\hfill
\begin{subfigure}[c]{0.48\linewidth}
    \centering
    \includegraphics[width=\linewidth]
    {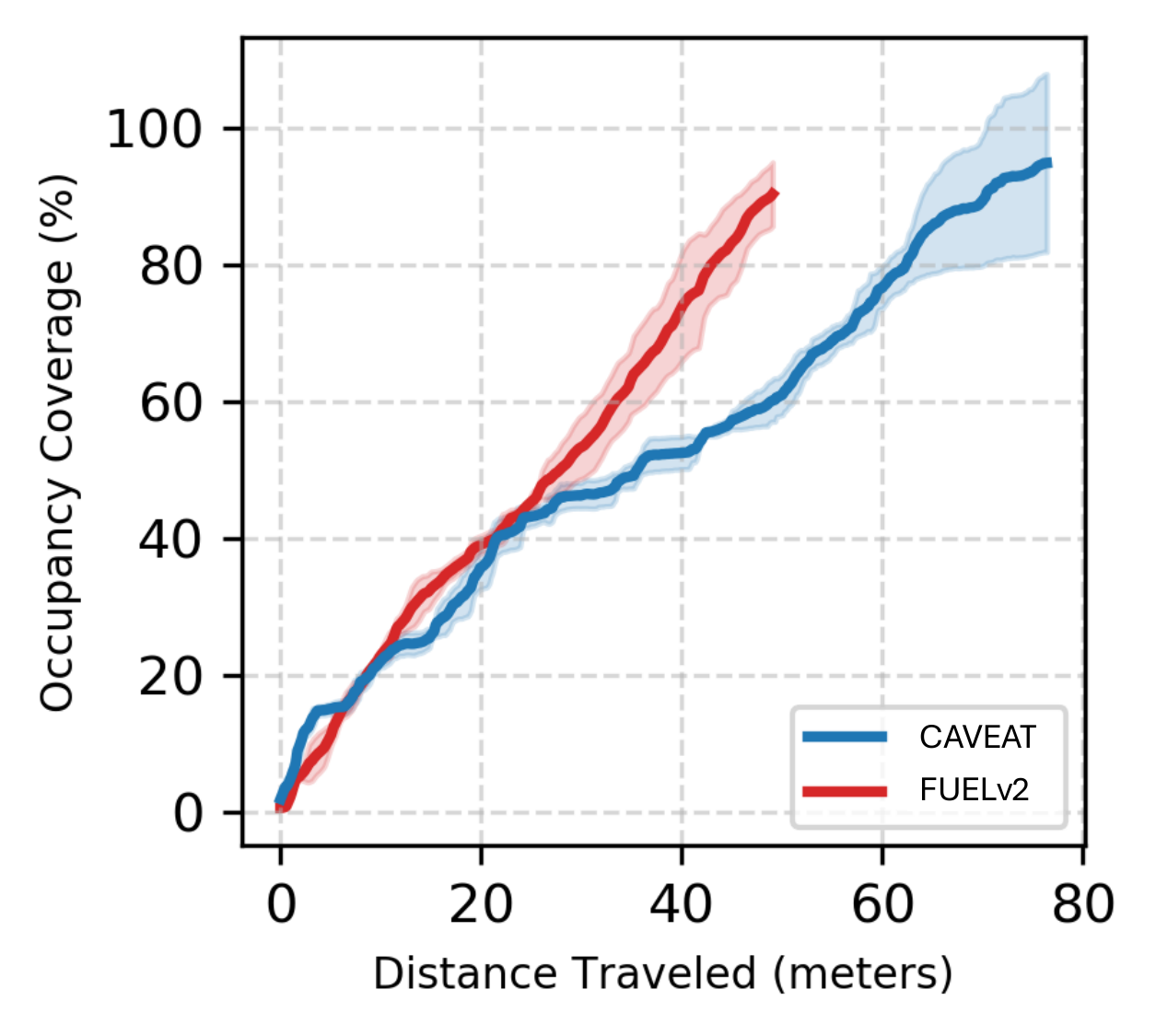}
    \caption{\ProjectName vs. FUELv2, env.~$B_{\mathrm{test}}$.}
\end{subfigure}
\caption{Occupancy coverage versus distance traveled in simulation. Panels (a)--(b) compare independent and rolling initialization; panels (c)--(d) compare SDF-guidance thresholds $d_{\min}=0.5\,\mathrm{m}$ and $d_{\min}=1.0\,\mathrm{m}$; panels (e)--(f) compare CAVEAT and FUELv2. Curves represent the average exploration trajectory across three to four runs; shaded regions represent $\pm 1$ standard deviation across the evaluation trajectories.}
\label{fig:simulation-comparisons}
\end{figure}

\subsubsection{Rolling-inference ablation}
\label{subsec:rolling_ablation}

The first ablation compares independent Gaussian initialization at each decision step with the rolling initialization of \secref{subsec:rolling_inference}. The observations, learned-policy parameters, diffusion settings, and local SDF guidance are unchanged.

As shown in~\figref{fig:simulation-comparisons}~(a--b), in environment $A_{\mathrm{test}}$, the variants initially attain similar coverage. Rolling inference subsequently reaches approximately $80\%$ coverage within $80\,\mathrm{m}$, while independent initialization remains near $60$--$65\%$ over a substantial additional distance and requires approximately $175\,\mathrm{m}$ to attain $95\%$. In environment $B_{\mathrm{test}}$, rolling inference reaches approximately $95\%$ coverage within $70\,\mathrm{m}$, whereas independent initialization attains approximately $85\%$ after $140\,\mathrm{m}$ and exhibits an extended plateau between approximately $70$ and $120\,\mathrm{m}$. Rolling initialization thus increases occupancy coverage per traveled distance in both tested environments.

\subsubsection{SDF-threshold sensitivity}
\label{subsec:sdf_ablation}

The second study compares the local SDF clearance thresholds $d_{\min}=0.5\,\mathrm{m}$ and $d_{\min}=1.0\,\mathrm{m}$. Rolling inference and all learned-policy parameters remain unchanged. 
The clearance threshold substantially affects coverage per traveled distance in environments containing narrow passages.

As shown in~\figref{fig:simulation-comparisons}~(c--d), in environment $A_{\mathrm{test}}$, the $0.5\,\mathrm{m}$ threshold reaches approximately $90\%$ coverage after $70\,\mathrm{m}$, whereas the $1.0\,\mathrm{m}$ threshold requires approximately $100\,\mathrm{m}$ to attain comparable coverage. In environment $B_{\mathrm{test}}$, the $0.5\,\mathrm{m}$ threshold reaches approximately $95\%$ coverage within $75\,\mathrm{m}$. With $d_{\min}=1.0\,\mathrm{m}$, coverage remains near $55$--$65\%$ over an extended interval and approaches completion after approximately $160$--$170\,\mathrm{m}$. The smaller threshold therefore provides greater occupancy coverage per traveled distance under the tested conditions, particularly in environment $B_{\mathrm{test}}$.

\begin{figure}[t]
    \centering
    \includegraphics[width=0.75\columnwidth]
    {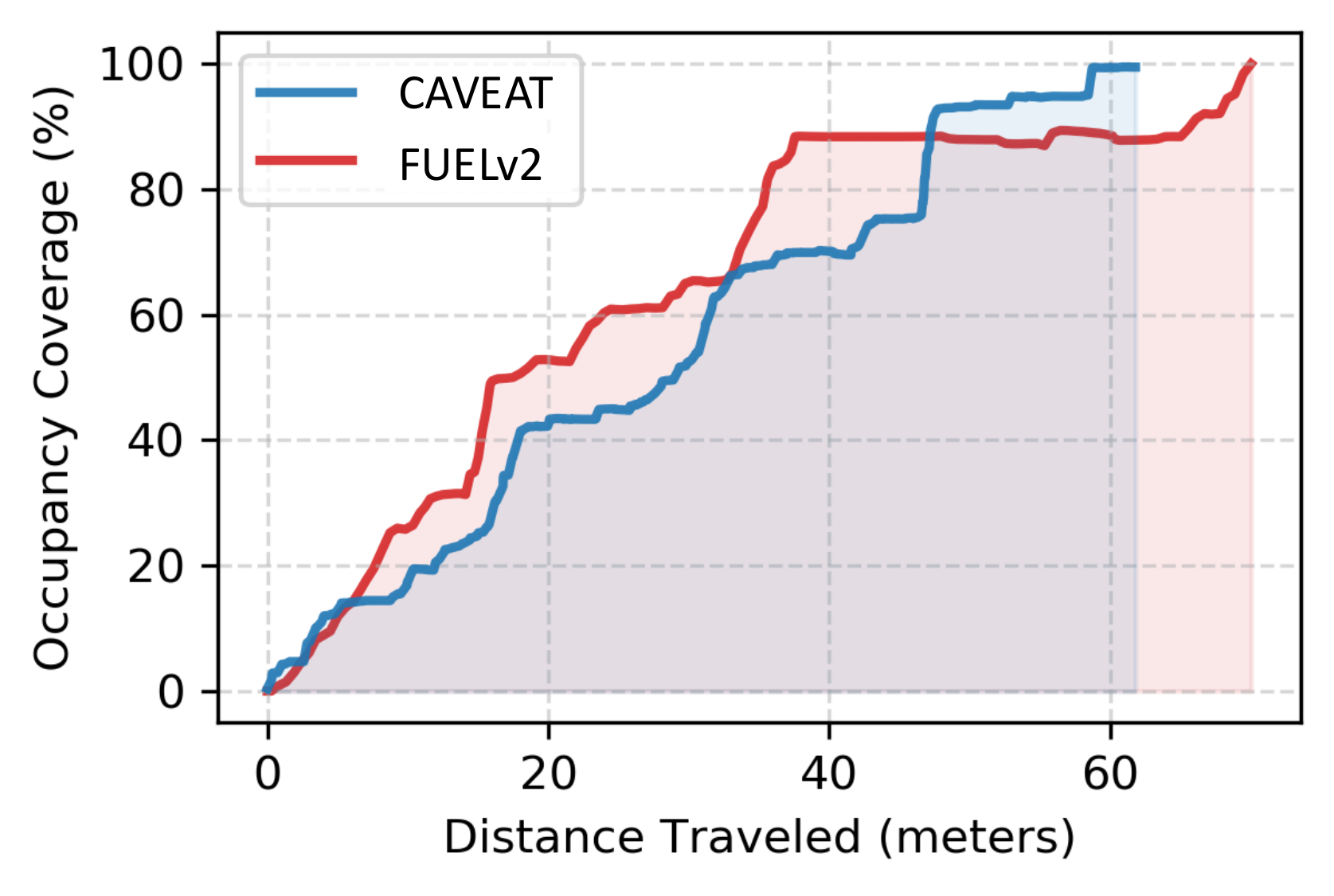}
    \caption{Comparison of \ProjectName and FUELv2, over the training environment $B_{\mathrm{train}}$.}
    \label{fig:caveat-vs-fuel-train}
\end{figure}

\subsubsection{Comparison with the FUELv2 expert}
\label{subsec:fuel_comparison}

Because \ProjectName is trained from FUELv2 demonstrations, this comparison evaluates the transfer of the expert's exploration behavior to a learned policy that does not execute the FUELv2 mapping and frontier-planning pipeline at deployment.

As shown in~\figref{fig:simulation-comparisons}~(e--f), in environment $A_{\mathrm{test}}$, FUELv2 reaches approximately $85$--$90\%$ coverage within $40\,\mathrm{m}$, while \ProjectName reaches approximately $90\%$ after $70\,\mathrm{m}$. In environment $B_{\mathrm{test}}$, FUELv2 reaches approximately $85$--$90\%$ within $45\,\mathrm{m}$, whereas \ProjectName attains approximately $60\%$ over the same distance and approaches $90\%$ after approximately $75\,\mathrm{m}$. \ProjectName therefore attains high final coverage in both environments, while FUELv2 provides greater coverage per traveled distance. To assess the role of the generalization capabilities required to explore a previously unseen environment, we have compared the two also on one of the training environments, namely $B_{\mathrm{train}}$. As shown in~\figref{fig:caveat-vs-fuel-train}, both \ProjectName and FUELv2 achieve complete coverage: \ProjectName over approximately $60\,\mathrm{m}$, and FUELv2 over approximately $65-70\,\mathrm{m}$. We leave as future work (i) improvement of the generalization capabilities of \ProjectName, and (ii) finer sampling of the training trajectories to improve the performance in more constrained regions, such as narrow passages and doorways, where it exhibits less efficient exploration behavior than FUELv2. 

\subsection{Real-World Proof of Concept}
\label{subsec:real_world}

The physical experiments evaluate exploration using DDPM and DDIM sampling and target-directed visual servoing using a separately trained instance of the same architecture.

\subsubsection{Exploration}
\label{subsec:real_world_exploration}

\begin{figure}[t]
\centering
\begin{subfigure}[c]{0.48\linewidth}
    \centering
    \includegraphics[width=\linewidth]
    {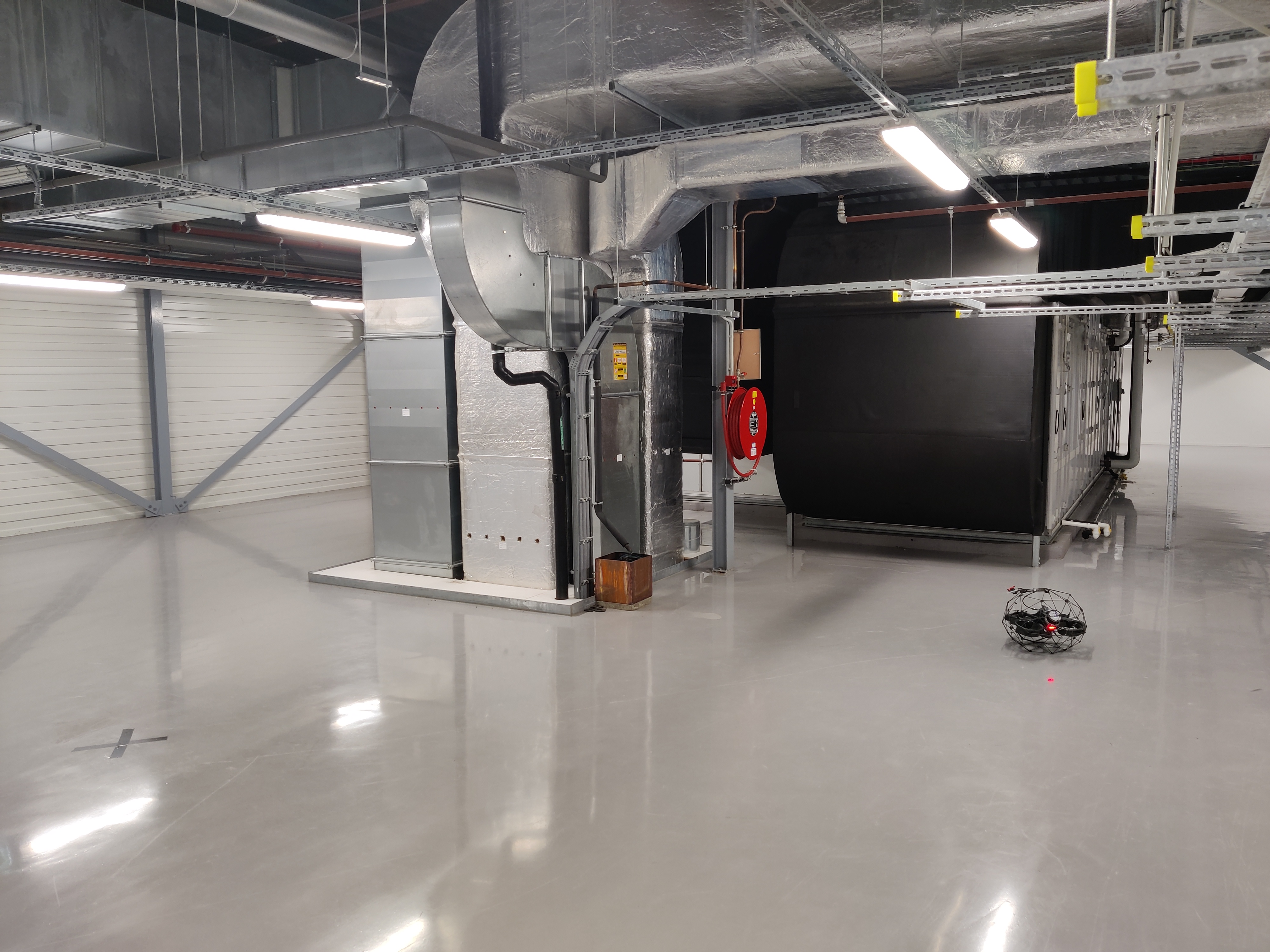}
    \caption{Open area and technical equipment.}
\end{subfigure}
\hfill
\begin{subfigure}[c]{0.48\linewidth}
    \centering
    \includegraphics[width=\linewidth]
    {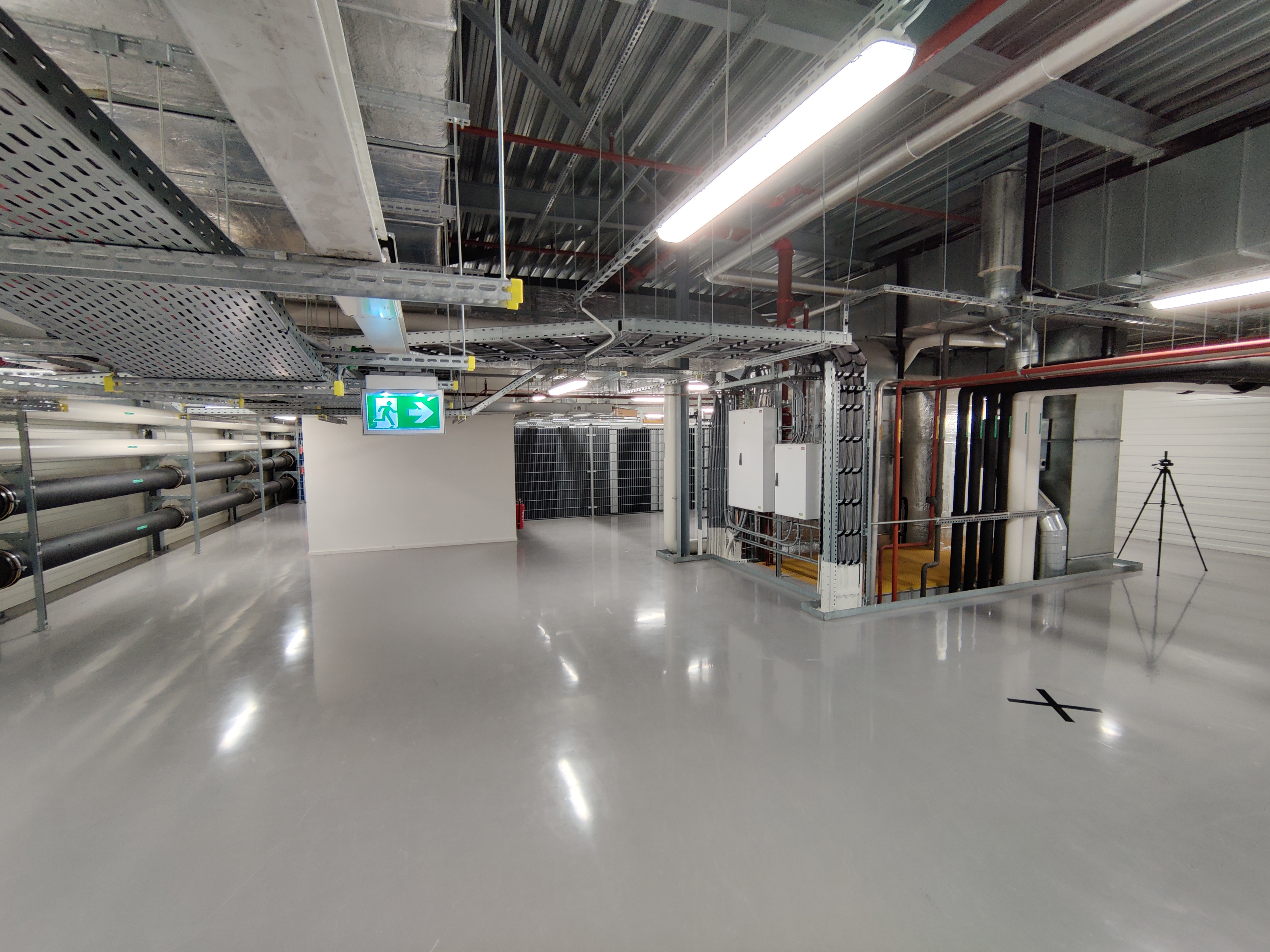}
    \caption{Pipes, installations, and passages.}
\end{subfigure}
\caption{Two views of the previously unseen real-world test environment.}
\label{fig:real-world-env}
\end{figure}

The exploration policy was deployed in the previously unseen indoor technical environment shown in \figref{fig:real-world-env}. The environment combines open floor areas with walls, ventilation ducts, pipes, structural columns, technical installations, and confined passages. The physical platform uses its onboard LiDAR, forward-facing camera, and proprietary pose estimate. The additional forward-facing LiDAR used in simulation is neither available nor required during physical deployment.

\begin{figure}[t]
\centering
\begin{subfigure}[c]{0.48\linewidth}
    \centering
    \includegraphics[width=\linewidth]
    {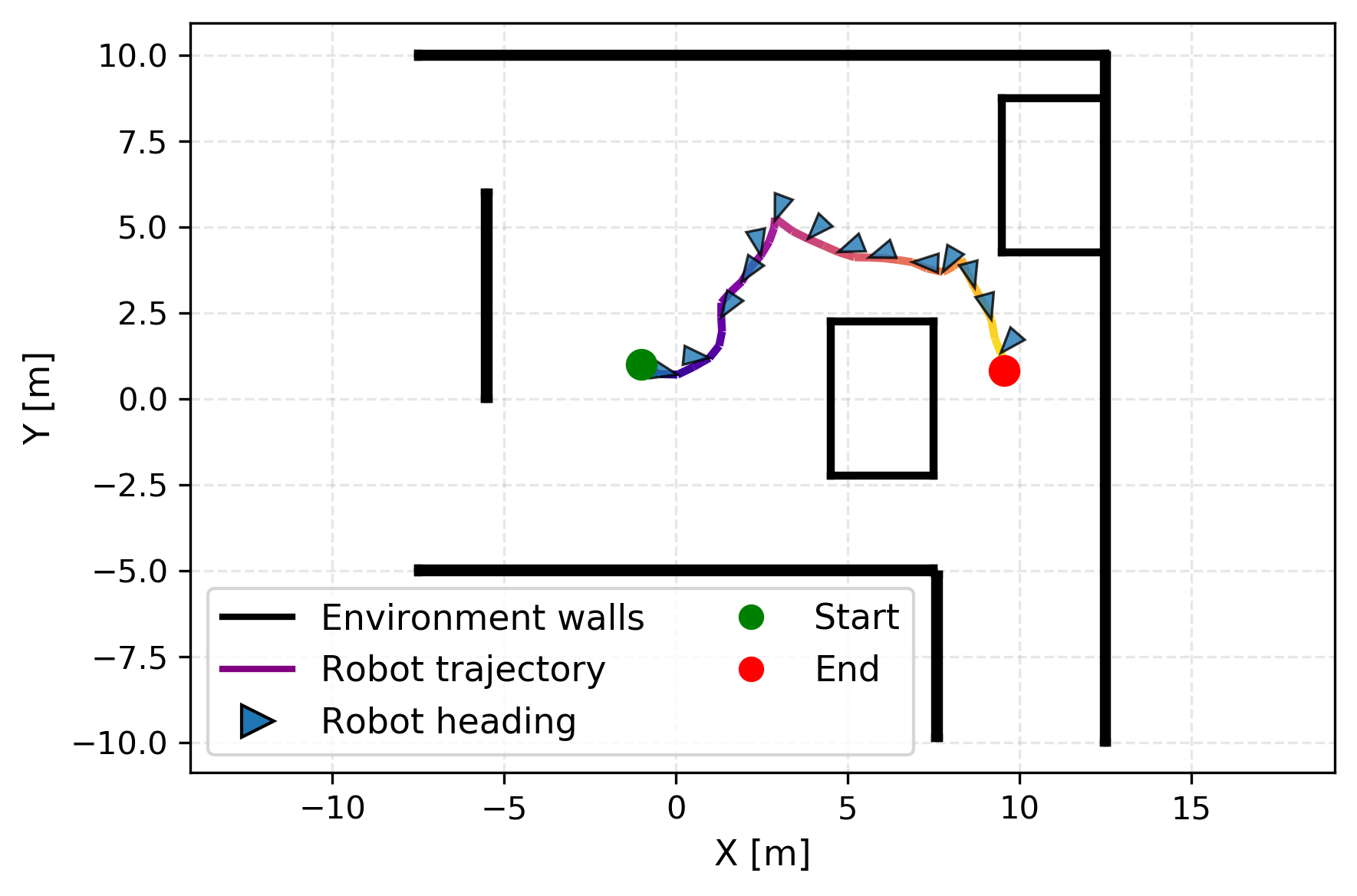}
    \caption{DDPM sampling.}
\end{subfigure}
\hfill
\begin{subfigure}[c]{0.48\linewidth}
    \centering
    \includegraphics[width=\linewidth]
    {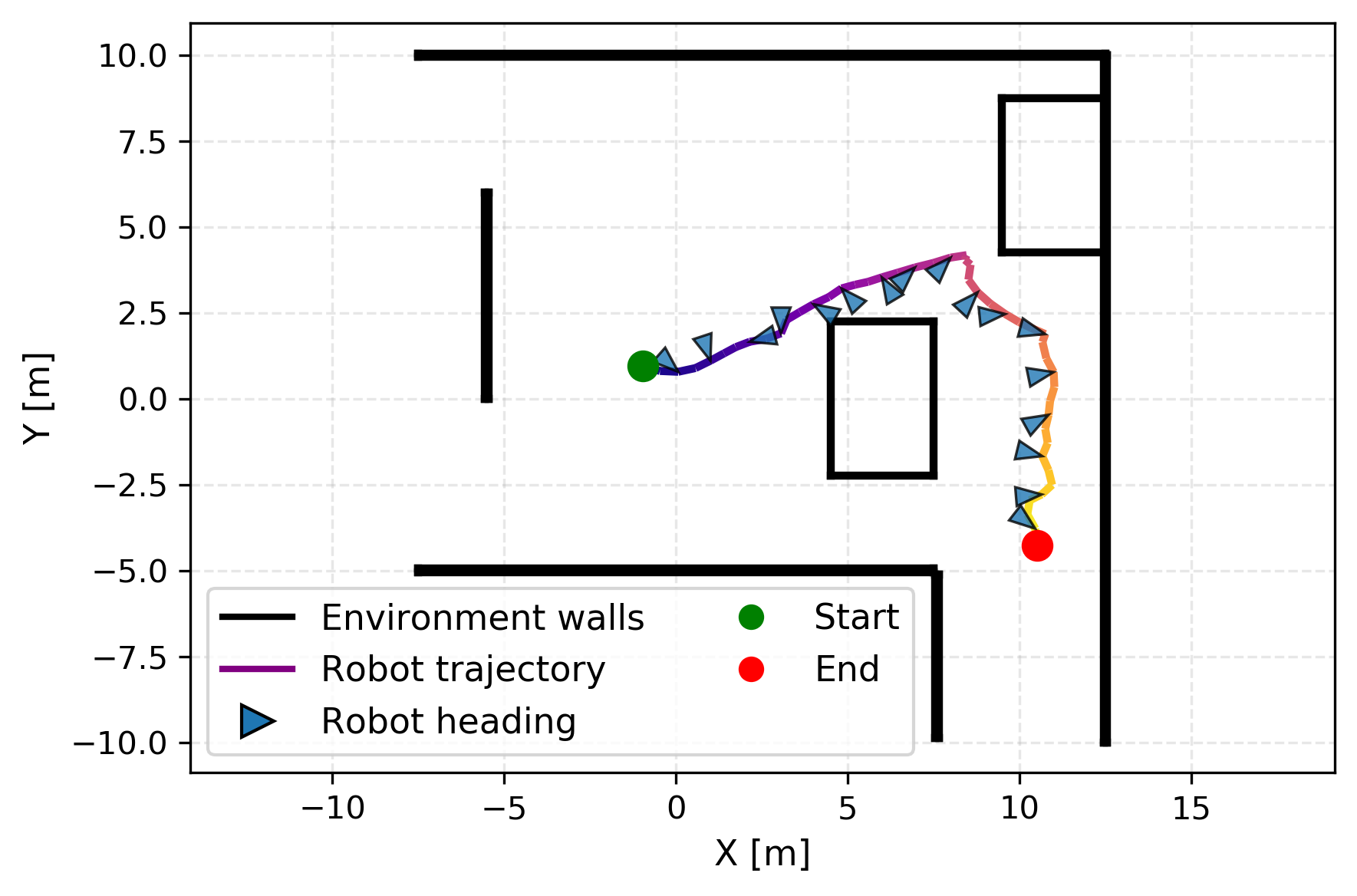}
    \caption{DDIM sampling.}
\end{subfigure}
\caption{Real-world exploration trajectories obtained with DDPM and DDIM sampling. Black lines indicate the principal environment boundaries, colored curves show the UAV trajectories, triangles indicate heading, and green and red markers denote the start and end positions.}
\label{fig:real-world-exploration}
\end{figure}

DDPM required 100 denoising steps and approximately $3.5\,\mathrm{s}$ to generate a waypoint sequence, whereas DDIM sampling \cite{ddim} reduced the number of steps to 10 and the average generation time to approximately $1.5\,\mathrm{s}$. The measurements were obtained on an off-board workstation equipped with an Intel Core i7-10750H CPU (6 cores/12 threads, up to 5.0 GHz) and an NVIDIA Quadro T1000 Max-Q GPU with 4 GB of VRAM, which receives the sensor data over WiFi as a ROS2 topic from the onboard controller. The commanded waypoints are also published as a ROS2 topic. \figref{fig:real-world-exploration} shows two representative individual flights.

Both samplers generated partial exploratory motion in the physical environment. In the displayed trials, the UAV progressed from the lower-left toward the upper-right portion of the mapped region while adapting its heading along the route. The shorter DDIM generation time enabled more frequent waypoint updates and produced the longer displayed trajectory within the available flight period.

The flights' duration was limited by the battery lifetime. Nonetheless, these trials demonstrate execution of the complete perception-to-waypoint pipeline and continued exploratory motion on the physical Elios~3.

\subsubsection{Target-Directed Visual Servoing}
\label{subsec:real_world_servoing}

A separately trained instance of the architecture was evaluated for target-directed visual servoing. The model was trained in simulation on trajectories approaching a predefined visual marker from randomized initial conditions. During initial physical trials, the policy could lose the target because the UAV heading did not consistently retain the marker within the camera field of view.

\begin{figure}[t]
    \centering
    \includegraphics[width=0.88\columnwidth]
    {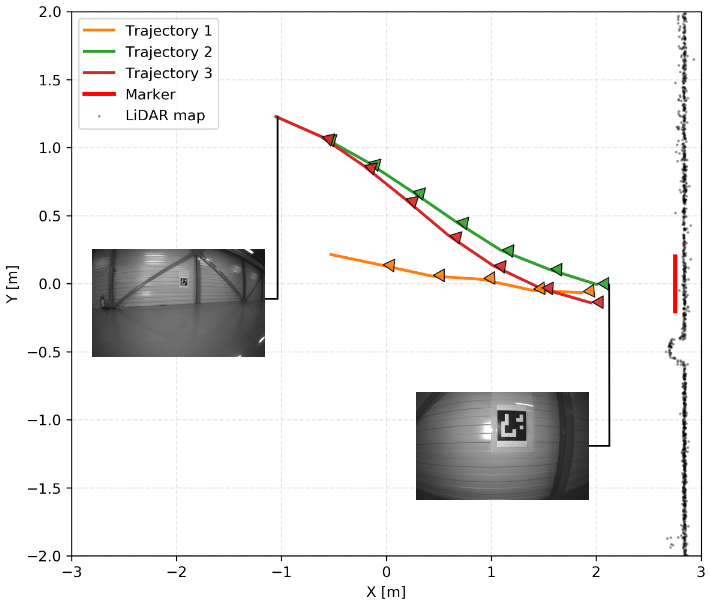}
    \caption{Target-directed visual-servoing proof of concept. Three trajectories approach the wall-mounted marker from different initial positions. Triangles indicate UAV heading, black points show the LiDAR-observed wall, and the inset images show representative camera observations.}
    \label{fig:real-world-servoing}
\end{figure}

The training images were subsequently augmented with Gaussian noise, motion blur, Gaussian blur, brightness and contrast variations, and radial lens distortion. With the augmented dataset, the policy generated the three target-approach trajectories shown in \figref{fig:real-world-servoing}. Starting from different positions, the UAV progressively oriented toward and approached the wall-mounted marker without the target-loss behavior observed before augmentation. An approach is considered successful when the target enters close view.

\subsection{Discussion}
\label{subsec:experimental_discussion}

The results support the feasibility of recurrent diffusion planning for simulated exploration and physical UAV deployment. They do not establish general equivalence to FUELv2, guaranteed collision avoidance, or broad real-world generalization. Exploration and target-directed visual servoing currently use separately trained policies. As future work, the model trained on a unified dataset can provide an integrated exploration-and-inspection paradigm.

\section{Conclusion}
\label{sec:conclusion}

This work investigated recurrently conditioned diffusion planning for aerial exploration without a persistent global map in the deployed policy. \ProjectName generates waypoint sequences from multimodal onboard observations and is trained from FUELv2 demonstrations. Simulations show that rolling inference improves coverage per traveled distance and characterize the effect of the local SDF clearance threshold. Physical experiments demonstrate partial exploration on an Elios~3 and target-directed visual servoing with a separately trained policy. Future work will study the information retained by the recurrent state, evaluate more diverse and geometrically distinct environments, and integrate exploration with target-directed behavior.

\bibliographystyle{IEEEtran}
\bibliography{bibliography}

\end{document}